\documentclass[referee,sn-vancouver]{sn-jnl}

\renewenvironment{table}[1][]%
{\tableorg[#1]%
\tablebodyfont%
\renewcommand\footnotetext[2][]{{\removelastskip\vskip3pt%
\let\tablebodyfont\tablefootnotefont%
\hskip0pt\if!##1!\else{\smash{$^{##1}$}}\fi##2\par}}%
}{\endtableorg}

\usepackage[utf8]{inputenc}
\usepackage{graphicx}
\usepackage[ruled]{algorithm2e}
\usepackage{hyperref}
\usepackage{verbatim}
\usepackage{amsmath,amssymb,amsfonts}
\usepackage{amsthm}
\usepackage{mathrsfs}
\usepackage[title]{appendix}
\usepackage{xcolor}
\usepackage{textcomp}
\usepackage{manyfoot}
\usepackage{booktabs}
\usepackage{listings}
\usepackage{rotating}
\usepackage{array}
\usepackage{multirow}
\usepackage{acro}
\usepackage{tabularx}
\usepackage{float}
\usepackage{pdflscape}
\usepackage{colortbl}
\usepackage{multirow}
\usepackage[table,xcdraw]{xcolor}

\newcommand{\fref}[1]{\footnote{\href{#1}{\nolinkurl{#1}}}}

\title{Language corpora for the Dutch medical domain}

\author[1,2]{\fnm{Bram} \sur{van Es}}\email{bes3@umcutrecht.nl}

\affil[1]{\orgdiv{Central Diagnostic Laboratory}, \orgname{University Medical Center Utrecht}, \orgaddress{\city{Utrecht}, \country{The Netherlands}}}
\affil[2]{\orgdiv{R\&D}, \orgname{B-lab}, \orgaddress{\city{Castricum}, \country{The Netherlands}}}

\begin{document}

\abstract{
\textbf{Background: }Dutch medical corpora are scarce, limiting NLP development.  \\
\textbf{Methods: }We translated English datasets, identified medical text in generic corpora, and extracted open Dutch medical resources.  \\
\textbf{Results:} The resulting corpus comprises $\pm$ 36 billion tokens across the medical domain in about 105 million documents, freely available on Hugging Face. \\
\textbf{Conclusion:} This work establishes the first large-scale Dutch medical language corpus for pre-training and downstream NLP tasks.
}


\maketitle

\section{Background}
\label{Intro}


Recent advances in natural language processing (NLP) have been driven by the availability of large, high-quality corpora such as C4, The Pile, and PubMedBERT’s training data \cite{Raffel2020,Gao2020,Gu2021}.  
However, for many minority and medium-resource languages, including Dutch, such domain-specific resources remain scarce.  
This is particularly true for the medical domain, where privacy concerns, copyright restrictions, and fragmented data sources make corpus creation challenging \cite{Neveol2018}.  

Several general-purpose Dutch corpora exist, such as COW \cite{Schafer2015}, OSCAR \cite{Ortizsuarez2020}, TwNC 
\cite{Ordelman2007} and SoNaR \cite{Oostdijk2008}. More recently large-scale web-crawled resources such as FineWeb2 and Finepdf \cite{Penedo2024} contain high quality Dutch corpora as does C4. 
While valuable for training large generic language models (e.g., RobBERT \cite{Delobelle2020}), these corpora contain little medical content and are not optimized for clinical or biomedical NLP.  
For clinical Dutch text specifically, resources are often limited to small annotated datasets for named entity recognition or classification tasks, typically derived from electronic health records under strict confidentiality agreements that require de-identification, see e.g. \cite{Menger2018} or \cite{Dernoncourt2017}.

As a result, researchers developing Dutch clinical NLP applications must either rely on translations of English corpora or manually curate small datasets. Both approaches have significant drawbacks: machine translations may introduce linguistic artifacts, and small corpora lack the coverage needed for robust model training.  

To bridge this gap, we present a large-scale Dutch medical corpus that combines multiple strategies: machine translation of existing biomedical corpora, automated identification of medical text in large Dutch web corpora, and targeted extraction of openly available resources such as PhD theses and professional guidelines.  
The resulting corpus, comprising approximately 35 billion tokens, is freely available and provides a broad, open resource for Dutch medical NLP research and model pre-training.

\section{Methodology}

\paragraph{Translation \& Transformation} of non-Dutch corpora.
We used a variety of decoder and encoder/encoder models to translate existing English corpora. 
Specifically NLLB by \cite{nllbteam2022} and MariaNMT by \cite{Tiedemann2020} for encoder/decoder models, and GPT3.5, GPT4o, Gemini1.5, Gemini2 and Gemini2.5 for the decoder models, and (accidentally) the Google Translate API\footnote{pro tip: don't start an unconstrained translation of GBs of text with a \$20/M-tokens model before you do your groceries}. 
We also translated several annotated corpora, including BioASQ and MedQA, using GPT4o-mini, and MIRIAD4.4 using GPT4.1-nano.

Besides \textit{direct} translations using LLMs we also transformed PMC Patient cases into structurally formatted discharge letters using Gemini 1.5 and Gemini 2.

We translate the Pubmed PMC archives using MariaNMT. For the machine translations we split the texts into non-overlapping, sentence delimited chunks of about half the maximum context length of the model, with half reserved for the decoder. For all the PMC content we have added the specific copyright statement belonging to each individual PMID.

\paragraph{Identification} of medical texts in generic Dutch corpora.
We used an LLM, GPT4.1-nano, to label $100.000$ random samples from the OSCAR dataset as medical/non-medical. We then trained a dense layer on top of a frozen RobBERT2023 encoder model on this labeled corpus. This classifier was then used to identify medical texts in FineWeb2 and FinePDFs. 

\paragraph{Extraction} of texts from open resources.
Via an Open Archives Initiatives\fref{https://www.openarchives.org/} (OAI)-connection with Dutch academic institutions we extracted Dutch PhD theses from which the Dutch/English content was extracted. For the selection we checked for medical keywords. We specifically selected right-free theses that were not under embargo. These PhD theses were parsed into an English medical corpus, as well as a parallel Dutch/English corpus based on the summaries and abstracts. The summary pairs were checked for multilingual similarity using sentence transformers (we used the \textit{paraphrase-multilingual-MiniLM-L12-v2} model). For the pdf-parsing we used PyPDF, Fitz and PDFminer in order of success. We checked whether the PDFs were the result of OCR by checking the producer in the Meta data and as a fallback we checked the number of pages that had text available in the page-objects\footnote{Quite arbitrary if more 75\% or more had no text, for a minimum of 15 pages, it was scanned, inversely if more than 75\% had text, for a minimum of 15 pages, it was not scanned}. If we decided that PDF was scanned we used PyTesseract to perform image-to-text on the list of images that we extracted using Fitz. 

We also share online resources such as NtvG publications and medical protocols from the federation of medical specialists (FMS) and the Dutch GP society (NHG).

\paragraph{Cleaning} the translations and the thesis-extraction were done using FTFY, and otherwise we applied regular expressions to remove spurious word repetitions and multiplicity of line breaks and spacing.

\paragraph{De-identification} was not performed on the invididual datasets, note that email and IP-address were already de-identified for the FineWeb2 and Finepdf corpora. We did perform a deidentification using the heuristics-based DEDUCE by \cite{Menger2018} on the combined set DutchMedicalTextV3\fref{https://huggingface.co/datasets/UMCU/DutchMedicalTextV3}.\\
For most of these tasks we used the (pre-alpha level) open source library PubScience\fref{https://github.com/bramiozo/PubScience}.

\section{Dataset characteristics}
%

\begin{table}[ht]
    \centering
    \begin{tabular}{ccccc}
    Source     & Extraction type & Words & Documents & Words/Document\\
         \hline
   FineWeb2 by \cite{Penedo2024}              & Identification & 11.5B& 11.7M & $\pm$ 500 \\
   Finepdfs by \cite{Penedo2024}              & Identification & 2.2B& 0.5M & $\pm$ 2000\\ 
   Pubmed Abstracts \fref{https://pubmed.ncbi.nlm.nih.gov/download/}            & Translation  & 2.6B & 15M & $\pm$ 200 \\
   Pubmed PMC OA Non-commercial\fref{https://pubmed.ncbi.nlm.nih.gov/download/} & Translation/Transformation & 8.7B & 3.4M & $\pm$ 2560 \\
   Pubmed PMC OA Commercial\fref{https://pubmed.ncbi.nlm.nih.gov/download/}     & Translation/Transformation & 4.4B & 1.5M & $\pm$ 2933 \\
   Apollo corpus by \cite{Wang2024}           & Translation    & 0.2B & 1.2M & $\pm$ 167 \\
   EMEA \fref{https://opus.nlpl.eu/EMEA/corpus/version/EMEA}   & Translation & 34M & 1662  & $\pm$ 36000\\
   MIMIC III/IV \cite{Johnson2016,Johnson2023} & Translation   & 2.9B & 9.4M & $\pm$ 308 \\
   ACGT  by \cite{Remy2023_2}                 & Translation    & 18M & 422K & $\pm$ 42 \\
   NtvG\fref{https://www.ntvg.nl/}            & N/A            & 34M & 44K & $\pm$ 773 \\
   FMS\fref{https://demedischspecialist.nl/}, NHG\fref{https://richtlijnen.nhg.org/}  & Scraping  & 200M & 364K & $\pm$ 550 \\
   other\footnote{PhD summaries, BioLORD \cite{Remy2022, Remy2023} , Meditron\cite{Chen2023} , etc.} &    &  2.2B    &  30M    &  $\pm$ 73.3         \\
   \hline
   total                &      & 35B & 73.5M &  $\pm$ 480\\ 
    \end{tabular}
    \caption{Overview of (most of the) datasets available with estimates of word counts, as of \textit{December 2025}. Note that these are the statistics for the datasets resulting from the translation effort.}
    \label{tab:placeholder}
\end{table}

\begin{table}[ht]
    \centering
    \begin{tabular}{cccc}
    Source                       & Extraction type           & Interaction type  & \# \\
         \hline
   MedicalFlashCards\fref{https://huggingface.co/datasets/medalpaca/medical_meadow_medical_flashcards}             & Translation               & Q/A open          &  32.9K \\
   MediQA by \cite{Asma2019}                       & Translation               & Q/A open          &  2.11K \\
   MedQA by \cite{Jin2021_2}                       & Translation               & Q/A MC            &  9.86K \\
   MedQA\footnote{The multiple choices are written out explicitly} & Translation & Q/A MC written    &  14.4K \\
   MedicalMMLU by \cite{Hendrycks2020}             & Translation               & Q/A MC            &  3.65K \\
   WikiDoc by \cite{Han2023}                       & Translation               & Q/A MC            &  5.76K \\
   SymptomDiseaseQA\fref{https://huggingface.co/datasets/prognosis/symptoms_disease_v1} & Translation & Q/A open & 10.1K\\
   BioASQ by \cite{Krithara2023}\fref{https://zenodo.org/records/7655130}            &     Translation  &   Q/A  & 4.7k  \\ 
   MIRIAD by \cite{Zheng2025}                          &  Translation              &   Q/A             &  4.4M \\ 
   \hline
   total                         &                           &                   &  54.28K \\ 
    \end{tabular}
    \caption{Overview of the interaction finetuning set available, as of \textit{December 2025}}
    \label{tab:placeholder}
\end{table}

\section{Models}
We trained several models on (parts of) this data. Specifically, as of writing;
\begin{itemize}
    \item CardioLlama.nl: domain-adapted pre-training of English Llama 3.2, 1B parameters,
    \item CardioBERTa.nl: continued pre-training of MedRoBERTa.nl, RoBERTa-based, 120M parameters,
    \item CardioDeBERTa.nl: from scratch training of DeBERTaV2, 400M parameters,
    \item MedLlama.nl: : domain-adapted pre-training of English Llama 3.2, 1B parameters,
\end{itemize}

and for future work we will train large multilingual models.

\section{Discussion}

This work presents the creation of a large Dutch medical language corpus that can be used for language model pre-training. Future work is the 
\begin{itemize}
    \item creation of more corpora for model finetuning and,
    \item the extraction and translation of \textbf{more} data.
\end{itemize}

This work is entirely re-producible and can be applied to minority languages \textit{other than Dutch}, contingent on the quality of machine translations and the availability of OAI-access to academic institutions. If more data or models are added, this document is updated.

\paragraph{Caveat:} the quality of the translated extractions cannot be assumed to be equal to that of the original texts, \textbf{regardless} of the model pedigree. All translation models merely approximate the intended representation of the original text in another language. 

\section*{Resources}
To perform this work a combination of resources was employed; workstations with a RTX2080, a RTX4000 ADA, a RTX4000 Quadro, two A10 GPUs and a v4-32 TPU pod, besides the LLM APIs.

\section*{Declarations}

\subsection*{Ethics approval and consent to participate}
The University Medical Center Utrecht (UMCU) quality assurance research officer confirmed under project number 22U-0292 that this study does not fall under the scope of the Dutch Medical Research Involving Human Subjects Act (WMO) and therefore does not require approval from an accredited medical ethics committee. The study was performed compliant with local legislation and regulations. All patient data were de-identified in compliance with the European Union General Data Protection Regulation, and as a result, written informed consent was not required by the UMCU ethical committee.

\subsection*{Availability of data and materials}
The datasets generated are freely accessible through Huggingface\fref{https://huggingface.co/UMCU/datasets}, the
codes used can be found on github\fref{https://github.com/bramiozo/PubScience},\fref{https://github.com/UPOD-datascience/TPU_compute}.

\subsection*{Competing interests}
The authors declare that they have no competing interests.

\subsection*{Acknowledgements}

The work received funding from the European Union's Horizon Europe research
and innovation programme under Grant Agreement No. 101057849 (DataTools4Heart project). \\
A substantial part of the translations have been performed on GPUs from SURF-SARA under projects EINF-15564 and EINF-11407.

\bibliography{bibliography} 

\end{document}